\documentclass{article}

\usepackage{arxiv}
\usepackage[utf8]{inputenc}
\usepackage[T1]{fontenc}
\usepackage{hyperref}
\usepackage{url}
\usepackage{booktabs}
\usepackage{amsfonts}
\usepackage{nicefrac}
\usepackage{microtype}
\usepackage{lipsum}
\usepackage{graphicx}
\usepackage{amsmath}
\usepackage[most]{tcolorbox}
\usepackage{xcolor}
\usepackage{tabularx}
\usepackage{booktabs}
\usepackage{colortbl}
\usepackage[numbers]{natbib}

\title{AISA: AI Safety Assistant Framework for Continuous Improvement of Highway Construction}

\author{
Mason Smetana \\
Department of Civil and Environmental Engineering\\
University of Pittsburgh\\
Pittsburgh, PA 15213 \\
\texttt{mrs196@pitt.edu} \\
\And
Trevor Neece \\
Department of Civil and Environmental Engineering\\
University of Pittsburgh\\
Pittsburgh, PA 15213 \\
\texttt{trevor.neece@pitt.edu} \\
\And
Lev Khazanovich \\
Department of Civil and Environmental Engineering\\
University of Pittsburgh\\
Pittsburgh, PA 15213 \\
\texttt{lev.k@pitt.edu} \\
}

\begin{document}
\maketitle
\begin{abstract}
\textbf{Objectives}: Job Safety Analysis (JSA) and pre-task planning can benefit from prior incident records, yet historical accident data is often stored as unstructured narratives that are difficult to consult at the point of planning. A novel framework centered on large language models (LLMs) for highway construction safety reporting and planning is proposed as a foundation for future agentic applications, prioritizing deterministic, local inferencing. The first aim is to enable classification and quality scoring of incident narratives for existing and future reporting purposes. The second is to evaluate retrieval of relevant historical accidents, related imagery, and trusted industry documents for incorporation into daily safety plans. 

\textbf{Methods}: Neural probes were trained to classify incidents along four multiclass and two binary Occupational Injury and Illness Classification System (OIICS) fields and to derive an overall quality score, evaluated on a test set of over 15,000 narratives and a held-out set of 100 author-labeled records, benchmarked against a majority-vote LLM ensemble. The retrieval of historical accidents, reference imagery, and industry documents was benchmarked across embedding models using standard information retrieval metrics.

\textbf{Findings}: OIICS classification reached 75\% held-out accuracy, though the two binary flags were degenerate. The quality score, while meaningful on one database, was distorted on out-of-distribution fatalities in the held-out dataset. Accident retrieval recovered relevant incidents far above chance, performing best on lexically distinct construction activities. On document question answering, an open-weight decoder embedding model surpassed proprietary models.

\textbf{Novelty}: Overall, this work provides a new framework rooted in local inferencing and text embedding models for future agentic applications, with emphasis on bridging external data to JSA reports.

\textbf{Practical Applications}: The findings guide practitioners toward affordable, auditable, privacy-preserving safety tooling that runs locally and folds new incidents and near-misses back into planning, without transmitting sensitive records to cloud providers.
\end{abstract}

\clearpage

\section{Introduction}
Highway and heavy civil construction work concentrates some of the most severe hazards in the American workplace \cite{Alm24}. In just a few examples, workers operate near heavy equipment, erect and dismantle formwork, and perform maintenance in close proximity to live traffic. The primary defense against these hazards is not any technology alone, but a practice of Job Safety Analysis (JSA) and pre-task planning, in which a safety professional breaks an upcoming task into steps, identifies hazards at each step, and selects controls before work begins \cite{Kow24}. The quality of said analysis depends almost entirely on whether the planner foresees potential hazards for a given work activity. Historical incident records, on one hand, provide the necessary foresight for dangers presented in similar work activities \cite{Bay26}; yet such repositories may be difficult to use or consider during the planning phase. They exist as thousands of unstructured accident narratives (free-text descriptions embedded within otherwise structured records), indexed by report metadata (e.g., dates, establishments, injury codes). Thus, connecting JSA to historical data remains a largely manual and therefore neglected step. At the same time, budget constraints and organizational resistance to change compromise the adoption of automation in construction, underscoring the need for approaches that are financially and operationally realistic rather than aspirational \cite{Mus24}. A tool that makes historical records instantly available during planning, without adding cost, directly serves this need.

Efforts to bring computation to safety have significantly progressed over the decade. Researchers have centered on predictive analytics, in which algorithms analyze historical accident reports alongside project variables such as weather, time of day, and site activity to forecast hazards and allocate safety resources \cite{Tix16}. Due to the recent advancements in natural language processing (NLP), specifically large language models (LLMs), attention has turned from predicting accidents to reading how they are reported. Deep NLP models have been used to classify construction safety narratives such as near-miss reports in support of hazard identification \cite{fang_automated-text_2020}, and related approaches have categorized inspection reports and inferred risk levels from free-text records \cite{li_automated-text_2025}. Moreover, \cite{yoo_harnessing-generative_2024} fine-tuned an LLM to predict accident types from safety narratives, reaching 82\% accuracy. However, a recurring caveat accompanies these advances, as fine-tuning and inferencing LLMs is computationally expensive and demands substantial labeled data, a requirement that is not frequently available in civil engineering contexts.

A parallel, recent line of work has produced conversational safety assistants or agents for personnel to use in daily operations. For example, TrafficSafetyGPT is an open-weight model fine-tuned on a specialized transportation safety dataset, extracts information from accident reports, and outperforms general-purpose models \cite{zheng_trafficsafetygpt-tuning_2023}. Similarly, AutoRepo advances automated inspection report generation, but is primarily image-driven and does not draw on historical accident data \cite{pu_autorepo-a_2024}. More recent systems, such as the Jin Tetsu assistant, are promising but explicitly limited by their dependence on cloud-based LLMs and associated high compute resources \cite{Cla26}. Taken together, the state of the art shares several gaps: none of these systems are specifically focused on highway construction; few integrate historical data into their reasoning; and most depend on cloud-based infrastructure. Requiring continuous connectivity in remote locations not only decreases usability but also leads to privacy concerns with potentially sensitive data, such as near-miss reports.

For these reasons, this paper proposes an AI Safety Assistant (AISA) framework for highway construction, with an emphasis on continuous improvement of JSA reports. The framework includes the use of historical accident repositories from publicly available Occupational Safety and Health (OSHA) databases, the use of related imagery, and the inclusion of trusted industry documents to produce more meaningful and specific JSA reports to assist daily planning efforts. The devised approach is intended to provide more deterministic, fast, and local tooling to be used either independently or in tandem with agentic LLM pipelines now entering practice. Two primary tasks are described and investigated in this study: (1) providing a classification model to report accidents into existing schemas, while also providing information about the quality of text-based narratives, and (2) pulling relevant historical data/imagery into JSA reports. The scoring mechanism proposed in the first task was designed to assist in the reporting of future accidents or near-misses, where in previous studies it was found that historical narratives tend to significantly vary in quality \cite{Sme26}. The belief is that in order for future NLP analyses to be meaningful, the reported narratives should contain enough information in the first place, with the ability to classify into a coded database without additional context.

For the second task, daily safety planning, a series of information retrieval tasks were carried out to inform future researchers and practitioners about the capabilities of modern text-embedding models -- the encoder portion of many LLMs \cite{Nee22}. The results provide information about how reliably incident narratives and relevant industry documents can be pulled into JSA reports with out-of-the-box embeddings through a simple Retrieval-Augmented Generation (RAG) pipeline \cite{lewis-retrieval-augmented_2020}. The emphasis on continuous improvement primarily lies between the interaction of existing JSA documents, stored accident data, and related imagery. Once a new incident or image is added to the system, a user may anticipate that it will be used in a future daily safety plan for a similar work activity -- such that it could better inform safety professionals with unintended hazards if they ever arise again. Notably, the data sources (e.g., historical accidents, images) described in this framework are strictly for training and validation purposes. The system is intentionally designed for proprietary and/or contractor-specific data to be incorporated.

\section{Methods}
\begin{figure}[ht]
\centering
\includegraphics[width=0.99\textwidth]{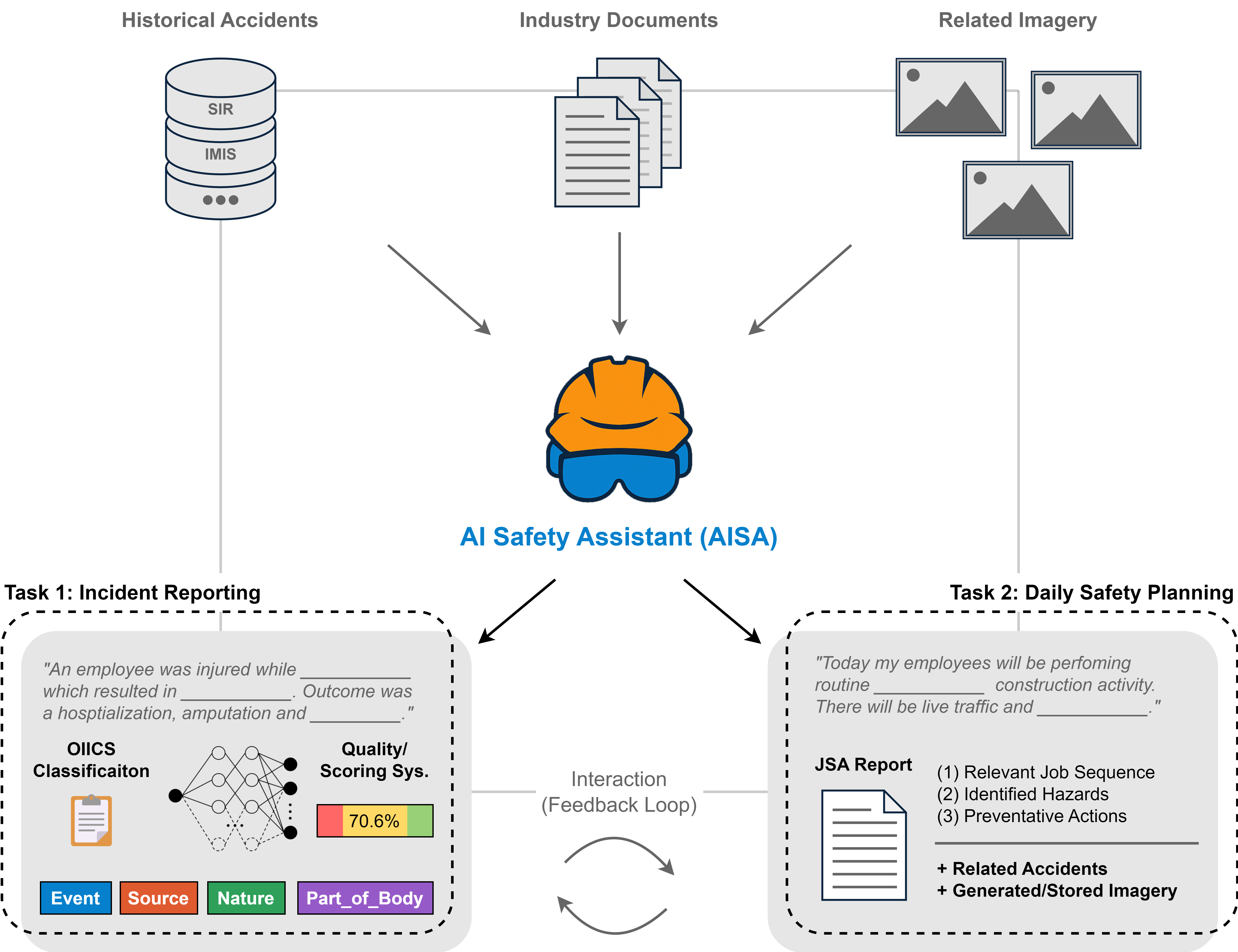}
\caption{Overview of AI Safety Assistant (AISA) Framework}
\label{fig:overview}
\end{figure}

The framework is organized around a central LLM safety assistant that draws on a common pool of construction safety information and applies it to two complementary tasks, as illustrated in \hyperref[fig:overview]{Figure \ref{fig:overview}}. Three shared data sources feed the assistant from the top: historical accident records from repositories such as historical accident databases, trusted industry documents (toolbox talks, standards, and specifications), and related imagery depicting equipment and site conditions. Together, this architecture positions itself not as a single model but as a coordinating layer that retrieves from, and reasons over, heterogeneous safety data.

The lower half of the figure shows the two tasks this framework supports. Task 1 processes unstructured accident narratives through a series of multilayer perceptron (MLP) probes to classify each narrative into a standardized coding structure and fuse their outputs into a single quality score to quantify a free-text narratives quality for reporting purposes. Task 2 works in the forward direction: given a description of the day's planned activity, the assistant uses RAG to assemble a JSA report (relevant job sequence, identified hazards, and preventative actions). This report is coupled with related historical accidents and stored or generated imagery. Critically, the two tasks are connected by a feedback loop, shown at the center of the figure. When a new accident, or near-miss, is reported in the field through the novel scoring system, it can be surfaced again in future daily safety planning activities. This interaction is what makes the framework one of continuous improvement rather than two isolated pipelines.

\subsection{Task 1: Incident Reporting and Quality Scoring System}
The first task of the AISA framework turns to the incident records themselves, with two core aims: (1) to classify incidents in a deterministic manner rather than relying solely on LLMs and (2) improve the quality of narrative reporting going forward, including for near-miss reports, which offer early insight into the precursors of serious incidents but are often recorded inconsistently \cite{fang_automated-text_2020}. Both goals depend on a single underlying capability: automatically judging how much usable safety information a given free-text narrative actually contains. The following subsections outline a text classification procedure to operationalize these goals, including what labeled data was available for training and immediate deliverables from the underlying architecture.

\subsubsection{OIICS Coding System}
The Occupational Injury and Illness Classification System (OIICS) is a standardized coding structure developed by the U.S. Bureau of Labor Statistics \cite{USD26}. It provides a consistent framework for classifying the specific characteristics of workplace injuries, illnesses, and fatalities. The system categorizes these incidents into four analytical perspectives: the nature of the condition (Nature), the part of the body affected (Part of Body), the source that produced the injury (Source), and the specific event or exposure that occurred (Event).

The OIICS hierarchy uses a 1-to-4 digit numerical system where adding digits narrows down the data from broad categories to hyper-specific details. The digits represent increasingly granular layers of classification: 1 Digit (Division): The broadest category, establishing the foundation of the classification (e.g., 2 for Transportation Incidents); 2 Digits (Major Group); Narrows the division down into an identifiable family (e.g. 24 for Pedestrian vehicular incidents); 3 Digits (Group): Adds a layer of specificity to the major group (e.g., 241 for Pedestrian struck by vehicle in work zone); 4 Digits (Subgroup / Terminal Code): The ultimate, most precise code possible in the system (e.g., 2411 for Pedestrian struck by vehicle propelled by another vehicle in work zone).

\subsubsection{Text Classification Pipeline}
A probe is a lightweight classification layer, here an MLP, trained on top of frozen embeddings to perform a target linguistic tasks \cite{perone_evaluation-of_2018}. Probing was introduced as an analytic tool: by holding the encoder fixed and training only a small classifier, one can examine how linguistic structure is represented in contextualized embeddings and how sentential context shapes word- and sentence-level tasks such as classification. The same lightweight setup is well suited to applied classification, and related ideas have seen use in civil engineering, with a particular focus on safety \cite{fang_automated-text_2020,li_automated-text_2025}. Comparable approaches have categorized construction inspection reports and inferred risk levels from free-text site records \cite{li_automated-text_2025}. In a related direction, Yoo et al. \cite{yoo_harnessing-generative_2024} fine-tune a GPT model to predict accident types from construction safety narratives, reaching 82\% accuracy; however, fine-tuning LLMs remains computationally expensive and requires substantial labeled data, which is often infeasible in civil engineering contexts. Probing a frozen embedding avoids both costs, since only a small classifier is trained.

In this study, the OSHA Sereve Injury Reports (SIR) database, as examined in \cite{sme24}, was used to train probing classifiers. Each record was already human coded to the OIICS framework and therefore provided labeled training data without additional annotation. While prior work has concentrated primarily on a single NAICS code (237310: Highway, Street, and Bridge Construction), comprising 1,032 entries from 2015 to 2021, this investigation uses the entire database across all industries from 2015 to 2025, over 100,000 entries, in order to learn patterns in a shared latent embedding space. OIICS version 2.0.1 is used instead of the modern version 3.0.1 since a majority of accidents were reported prior to 2023 when the system was updated. Six separate MLP networks were trained to predict one categorical field from the free-text incident narratives in the database, operating on frozen nomic-embed-text-v1.5 embeddings \cite{Nus24}. The fields are Event, Source, Nature, Part of Body (level 2 hierarchy), and binary indicators for Hospitalized and Amputation.

\begin{figure}[ht]
\centering
\includegraphics[width=0.99\textwidth]{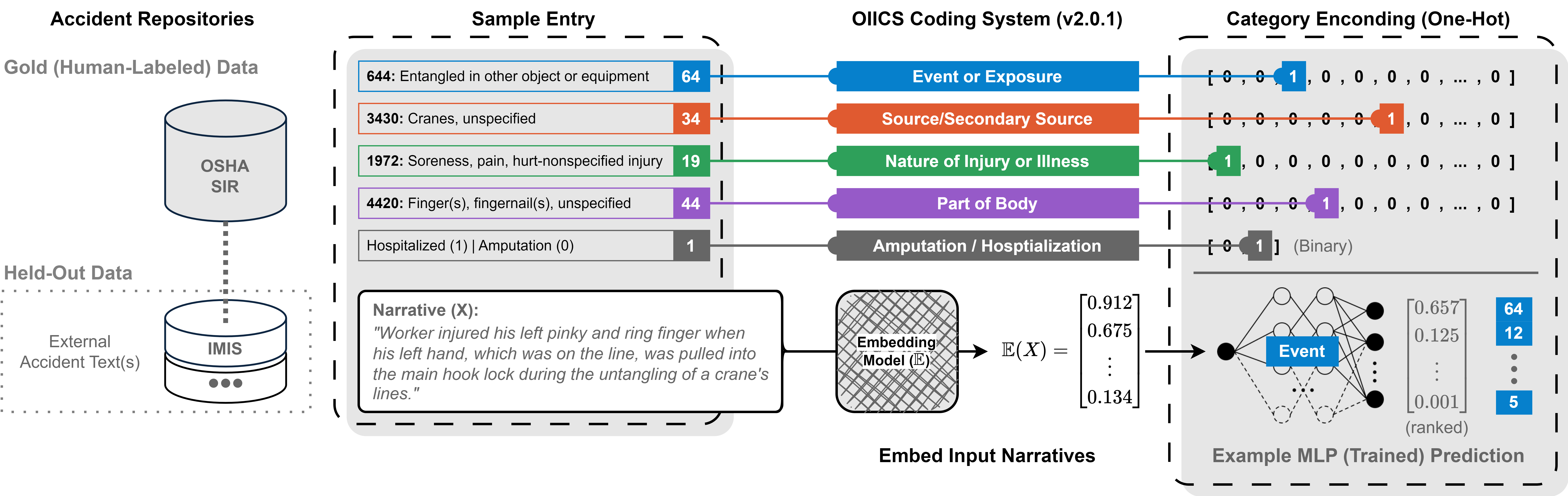}
\caption{Overview of MLP architecture for predicting OIICS accident codes}
\label{fig:prediction}
\end{figure}

\hyperref[fig:prediction]{Figure \ref{fig:prediction}} illustrates the classification pipeline. Accident records were drawn from the human labeled (``gold'') SIR database and held-out external narratives from a separate IMIS database, as investigated in \cite{Sme26}. Each record's OIICS codes were truncated to their two-digit major group (e.g., Event code 644 $\rightarrow$ 64, Source 3430 $\rightarrow$ 34) and converted to a one-hot target vector \cite{Dah21}, such that each network learns a probability distribution over all classes ($C$) within its category. The narrative text ($x_{i}$) is mapped by the frozen embedding model to a fixed dimensional ($d$) vector $e_{i}=\mathbb{E}(x_i)$, which is the shared input to all six MLPs. Each network then produces a ranked probability distribution over its major group classes; the top-ranked class (e.g., Event 64 in the figure) is the model's prediction.

The networks were designed as one-hidden-layer MLPs (hidden size 256, dropout 0.3, learning rate 0.001, batch size 128, weight decay 0.0, trained for up to 400 epochs), optimized with cross-entropy loss \cite{Guo17} under a 70/15/15 split. The process begins with the precomputed embedding vectors $e_{i}$. Each vector is then passed through a fully connected linear layer and followed by a non-linear activation $\phi(\cdot)=\text{ReLU}$, which yields the hidden representation $h$. The hidden layer output is then mapped by a second fully connected layer to a vector of raw scores, or logits, one per class. The softmax function transforms these logits into a probability distribution in which each value lies in $[0, 1]$ and all values sum to 1; the class with the highest probability is the model's prediction.

\begin{equation}
    \label{eq:mlp}
    h=\phi(W_1 e_i + b_1), \quad z=W_o h + b_o, \quad p=\text{softmax}(z)
\end{equation}

Where $W_1, b_1$ and $W_o, b_o$ are the weights and biases of the hidden and output layers, $z$ are the logits, and $p$ is the resulting class-probability distribution. Thus, the network $f$ produces a $p$ distribution over input text $x_i$, $p=f[\mathbb{E}(x_{i})]$. Formally, incident narratives were considered as a multiclass classification problem with $C$ classes (e.g., $C=49$ for the Event field, $C=2$ for hospitalization and amputation). Given labeled data from the SIR database, with $y_{i} \in \left\{1, \dots, C\right\}$, fixed embeddings $e_i \in \mathbb{R}^{d}$ ($d=768$ for nomic-embed-text-v1.5) were computed through nomic's pretrained encoder for each narrative -- following a common ``encode-then-classify'' evaluation protocol for sentence representations \cite{Con18}. Each network is trained with cross-entropy loss:

\begin{equation}
    \label{eq:xentropy}
    \mathcal{L}=- \log {p (y = y_i | x_i)}
\end{equation}

\subsubsection{Prediction-based Uncertainty Scoring}
A secondary goal was to fuse those per-column outputs into a single cumulative score that represents the quality of the original free-text input to how well the text lets the models populate the record. The central difficulty is that ``quality'' is not one quantity. A per-class confidence conflates at least three things that are better kept apart: how informative the text is (does it carry enough signal to fill the fields), how in-distribution it is (is it the kind of text the models ever saw), and how determinate each column is (some fields are inherently ambiguous even for perfect input).

For each network and each input, the pipeline computes four quantities (ignoring Hospitalization and Amputation Binary columns). The prediction was then formed by applying the softmax over the classes and taking the argmax and its probability (top-1). However, top-1 confidence looks only at the single largest probability and is blind to the shape of the rest \cite{Gaw22}. Two classes can both report a top value of 0.4 while meaning very different things: one splits the remaining mass between two plausible classes (a genuine near-tie, still informative), the other smears it across $C$ classes (real confusion). Entropy ($H$) captures this distinction because it summarizes the whole distribution rather than its peak, describing the average level of information carried by the MLP prediction; that whole distribution reading is what lets entropy serve as a pseudo-rating of narrative quality. Normalizing by the maximum possible entropy, $\log(C)$, makes columns with different class counts comparable. The determinacy was then computed as one minus that normalized entropy \cite{Pas25}:

\begin{equation}
    \label{eq:determinacy}
    H(p)=-\sum_{i} p_i \cdot log(p_i), \quad det = 1 - H(p)/ \log(C)
\end{equation}

It is important to note that entropy measures concentration and uncertainty, not correctness. A model can be confidently wrong -- all mass on one wrong class gives $H_{\text{norm}} \approx 0$, which looks like high quality. Low entropy essentially means ``the model has committed,'' not ``the model is right.'' The final pseudo-quality score is taken as the average of determinacy across MLP predictions and is still a preliminary system that should be evaluated with greater scrutiny in future studies.

\subsubsection{Interpretability and Word Saliency}
To identify which words in the input narrative influence a probe's prediction, a word saliency analysis was implemented. Neural networks otherwise remain black-box in nature \cite{yoo_harnessing-generative_2024}, and occlusion-based saliency addresses this by measure how the prediction changes when parts of the input are removed. The single-word erasure was computed using a leave-one-out (LOO) perturbation strategy \cite{li_understanding-neural_2017}. For this calculation, let the input narrative be considered as sequence of words or tokens $x_{i} = \left\{w_1, w_2, \dots, w_n\right\}$ and $k$ be the index of the maximum probability from the underlying softmax distribution (corresponding to class $k$ in $C$ total classes). The importance score ($IS_{j}^{\text{LOO}}$) for word $w_j$ is calculated as follows:

\begin{equation}
    \label{eq:loo}
    IS_{j}^{\text{LOO}} = p_k(x_i) - p_k(x_i - w_j), \quad x_i = \left\{w_1, w_2, \dots, w_n\right\}
\end{equation}

A large positive importance means removing a word from the original input ($w_j$) collapsed the model's confidence in $k$, so $w_j$ supported the prediction; a negative value means the word was pushing the model away from $k$. Scores near zero mark tokens the prediction did not depend on. This operator answers ``which parts of the input support the top-1 prediction?,'' and is purely for exploratory purposes and interpretability of the resulting OIICS predictions.

\subsection{Task 2: Daily Safety Planning and Retrieval}
The second task of the framework addresses daily safety planning by grounding the assistant in an established source of structured safety knowledge: the Pennsylvania Department of Transportation's JSA Manual, Publication 517 \cite{Pen23}. In prior work, \cite{Sme26} demonstrate that this document provides a robust ontology for reasoning about highway construction safety, using it to map unstructured OSHA accident narratives onto structured job/step/hazard categories for asphalt pavement construction. That study established that this manual is already extensive and inclusive: its 47 maintenance activities span the equipment, work sequences, and potential hazards of the pavement maintenance cycle, and a dual LLM classification ensemble was able to align a majority of real accidents to a corresponding job. Second, and equally important, that same analysis exposed the limits of relying on the JSA alone. A substantial share of accidents corresponded to activities absent from the taxonomy (e.g., milling, vibratory compaction). Daily safety planning therefore cannot be a static lookup against a fixed document; it requires a mechanism to supplement the JSA with evidence from real incidents and external references.

RAG provides exactly this mechanism, which supplies an LLM with relevant external information at inference time, rather than relying solely on the knowledge encoded in its fixed weights. Instead of modifying the model, RAG retrieves passages from a document corpus based on their relevance to the input query and injects them into the prompt, allowing the model to generate a more informed, factually grounded response \cite{lewis-retrieval-augmented_2020}. In another related study, the technique was applied to concrete pavement construction, vectorizing domain documents, ranking the most relevant passages, and injecting them into the prompt \cite{smetana_evaluating-expertise_2025}. This retrieval-based grounding is particularly attractive for AISA, which should stay current with evolving standards and site specific conditions: the corpus can be updated with new documents at any time, and different agencies or regions can substitute their own specifications without retraining.

In its simplest form, often termed naïve RAG, both the query ($q$) and each candidate document ($d$) are mapped to fixed dimensional embedding vectors by the same encoder ($\mathbb{E}$), and relevance is scored by the cosine similarity between them \cite{Nee22}:

\begin{equation}
    \label{eq:cosine}
    \cos (q, d) = \frac{q \cdot d}{\|q\| \|d\|}
\end{equation}

This study adopted the cosine similarity retrieval for its transparency and low computational cost. It is evident that modern variants exist, including agentic search \cite{LiZ26}, in which a model directly searches through a corpus of text instead of dense retrieval, and that such approaches may offer accuracy gains at the expense of added complexity. The following open-weight embedding models were included in analysis as well as OpenAI proprietary models' text-embedding-3-small and text-embedding-3-large \cite{Opery}: Qwen3-Embedding-0.6B: 600 million parameters, $d=1024$, 32k context size, decoder-only Transformer architecture \cite{Zha25}; nomic-embed-text-v1.5: 137 million parameters, $d=768$, context size 8192, encoder-only Transformer architecture \cite{Nus24}; all-MiniLM-L6-v2: 22 million parameters, $d=384$, context size 512, lightweight encoder-only distilled Transformer architecture \cite{Rei19}. The following metrics were used in RAG evaluation: Recall (R), Mean Average Precision (MAP), Mean Reciprocal Rank (MRR) and Normalized Discounted Cumulative Gain (NDCG) -- which are typical for information retrieval tasks \cite{Sal24}.

\subsubsection{Historical Accident Records}
The first retrieval stream draws on past incident narratives, from databases like SIR and IMIS, surfacing how similar work has led to injury in the field. For a given planned activity, the assistant retrieves the accident records most relevant to it, such that a crew's safety briefing is informed by real, comparable events rather than by generic hazard lists alone. Based on the results of the previous analysis of PennDOT Pub517, a total of 965 narratives could be appropriately mapped to the 46 of the 47 construction activities described in the JSA manual -- one class with zero labeled accident is excluded. To assess accident retrieval quality, each activity was described by a short free-text description, which served as query $q$ in Eq (\ref{eq:cosine}). For example, the ``25: Patching -- General Procedures'' activity was labeled as ``Repairing damaged areas of pavement by cutting or breaking out deteriorated sections and restoring the surface. Workers mark the pavement, remove debris, apply a tack coat, place patching material, and compact it using rollers or hand tampers.'' The corpus to search against was the 965 labeled accidents (according to which activity they belong to).

\subsubsection{Related Imagery}
Retrieval extends to visual references associated with the activity or its equipment, giving crews concrete depictions of hazardous configurations rather than text descriptions alone. It has been shown that multimedia communication of information, in which it is presented both through multiple delivery methods (e.g. text, audio, images, etc.), is retained more readily than information presented with just one \cite{Pai86}. This relationship holds true for the construction industry, in which visual-based storytelling helps workers recognize warning signs before a hazard peeks \cite{Cul08}.

To enable the potential benefits of multimedia communication within JSA documents, this framework leverages generative AI to visualize highway construction hazards using synthetic images. This process was presented by Neece et al. \cite{Nee26}, and produces new images accompanying records form the SIR database under NAICS code 237310, which was already described previously. Each narrative was fed into a text generation model to convert it into structured descriptions of three layers: (1) infrastructure (the base roadway or work zone layout), (2) activity (the work being performed with equipment and workers), and (3) hazard (the peak worker-hazard interaction). These descriptions were then passed into Google's Nano Banana image generation model \cite{Goo25}. A visual constraints prompt, specifying a photorealistic, eye-level locked perspective, wraps the structured narrative description and generates the image in one pass. The images, however, were retrieved with their linked text-based narrative from the accident database in the same manner as the prior. This utility is purely to provide the AISA prototype and daily safety planning tasks with visual aids.

\subsubsection{Industry Document Question Answering}
Finally, the system searches a curated set of over 200 vetted external references: toolbox talks, agency standards, and equipment documentation -- sourced from publicly available repositories with no conflicting commercial interest. The purpose of these documents was to surface supplementary controls and procedures beyond what typical JSA documents enumerate. This mirrors the document injection approach validated for concrete pavement construction \cite{smetana_evaluating-expertise_2025}, in which source documents augment task-level hazard reasoning rather than replacing it. Examples of these resources include worker-facing safety guidance from The Center for Construction Research and Training (CPWR), such as the ``Prevent Exposure: Silica Dust in Enclosed Cabs'' toolbox talk \cite{CPW20}, and state agency documents including PennDOT's Daily Safety Talk Book \cite{Pen22}.

To convert this unstructured corpus into supervised retrieval data without manual annotation, the approach described by NVIDIA for synthetic data generation was followed \cite{NVI26}. The pipeline used an LLM, gpt-4o-mini \cite{Ope241}, to read the documents and automatically generate (query, relevant-document) question-answer (QA) pairs, avoiding hand labeling that is expensive, slow, and biased by the annotator's interpretation of relevance. The adopted stages are as follows: (1) segment documents into passages that are suitable, (2) generate artifacts and question-answer pairs, and (3) evaluate using LLM-as-judge, as suggested in \cite{Zhe23}. This ultimately produced a labeled corpus of over 6,000 questions and their supporting passages to provide a retrieval benchmark.

\section{Results}
\subsection{Incident Classification and Scoring}
\subsubsection{OIICS Classification Models}
\begin{figure}[ht]
\centering
\includegraphics[width=0.75\textwidth]{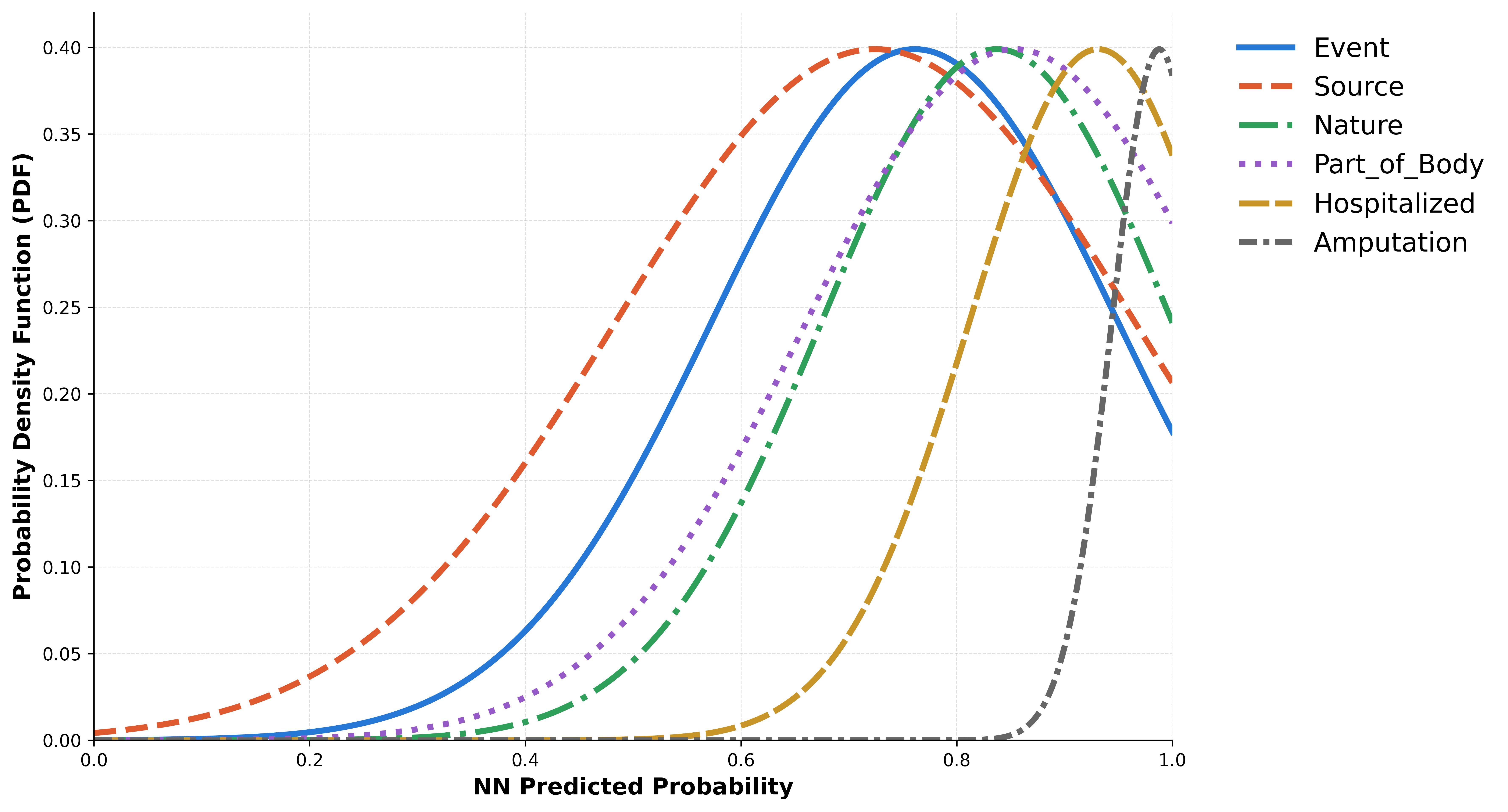}
\caption{Probability density function plotted for each OIICS classification model on SIR test set}
\label{fig:oiics}
\end{figure}

The historic SIR database of 102,922 incidents was distributed over a 70/15/15 training, validation, and test split. Each free-text narrative was first embedded by nomic-embed-text-v1.5 \cite{Nus24}, and classified alongside the OIICS, for which separate probe MLPs were trained by code structure over the same split. The four multiclass categorical variables include: Event (No. Classes, $C=49$), Source ($C=79$), Nature ($C=41$), and Part of Body ($C=46$). Binary variables ($C=2$) include Hospitalized and Amputation where labels can either be 1 or 0. During training, the models typically converged at approximately 50 epochs on average before the validation datasets exhibited overfitting and validation loss began to inflect in the positive direction (lower is better for cross-entropy). The trained models tend perform well on the held-out SIR test set ($n = 15,439$), but performance is strongly and interpretably segregated by task difficulty: the binary flags are near perfect, the multiclass structures are strong, and the single weakest predictive capability is for the Source code. The two views in \hyperref[fig:oiics]{Figure \ref{fig:oiics}} and \hyperref[tab:oiics]{Table \ref{tab:oiics}} are consistent where the shape of a model’s confidence distribution mirrors its measured accuracy.

\hyperref[fig:oiics]{Figure \ref{fig:oiics}} fits a normal distribution to each model's predicted probability across the test set, all skewed left meaning they lean toward high probabilities. The two features that carry meaning are their centers (the model's mean confidence $p$) and their width $\sigma$ (how consistent that confidence is). A curve pushed hard against 1.0 indicates a model that routinely commits to a single code; a broad curve sitting at lower probability indicates a model that frequently spreads its mass across several plausible indices rather than settling on one. Amputation ($\bar{p}=0.99, \sigma =0.04$) is a tall, narrow spike jammed against 1.0 -- it almost always predicts decisively. Source ($\bar{p}=0.72, \sigma =0.24$) is the broadest and flattest curve in the figure, the signature of a model that is often less confident in its prediction.

Accuracy in \hyperref[tab:oiics]{Table \ref{tab:oiics}} describes model correctness against the original OIICS labels. Acc@1 is top-1 accuracy, or the fraction of narratives for which the model's highest probability code matches the gold label. Similarity, Acc@3 relaxes this to whether the gold code appears among the model's top three prediction. The remaining metrics, NDCG@10 and MRR@10, summarize the quality of the full predicted ranking (higher is better). Reading across the test set, the numbers track difficulty cleanly. Part of Body is the strongest multiclass model (Acc@1 = 0.843), with Nature close behind (Acc@1 = 0.815) -- both of which pair high accuracy with tight, high-centered confidence distributions ($\sigma \approx 0.16 - 0.19$). Event performs decently over its 49 classes (Acc@1 = 0.753). Source (Acc@1 = 0.694) is the weakest across every statistic, which is unsurprising given it must resolve the largest and most overlapping taxonomy (79 codes). The binary flags, Hospitalized (Acc@1 = 0.931) and Amputation (Acc@1 = 0.989) are effectively solved. Since only two possible classes can be predicted, the remaining metrics are not sensible and are excluded from evaluation. It is worth mentioning that this level of MLP complexity may not be necessary for binary classification, and simpler techniques like logistic regression may be more than sufficient.

\begin{table}[ht]
\centering
\caption{OIICS classification performance summary on SIR test set ($n=15,439$)}
\label{tab:oiics}
\begin{tabular*}{\textwidth}{@{\extracolsep{\fill}} l *{6}{c} @{}}
\hline
\textbf{OIICS Model} & \textbf{No. Classes ($C$)} & \textbf{Avg. Prob ($\bar{p}$)} & \textbf{Acc@1} & \textbf{Acc@3} & \textbf{NDCG@10} & \textbf{MRR@10} \\
\hline
Event & 49 & $0.76\pm0.19$ & 0.753 & 0.949 & 0.889 & 0.854 \\
Source & 79 & $0.72\pm0.24$ & 0.694 & 0.876 & 0.838 & 0.795 \\
Nature & 41 & $0.84\pm0.16$ & 0.815 & \textbf{0.978} & 0.923 & 0.896 \\
Part of Body & 46 & $0.85\pm0.19$ & \textbf{0.843} & 0.960 & \textbf{0.927} & \textbf{0.904} \\
Hospitalized (Binary) & 2 & $0.93\pm0.12$ & 0.931 & n.a. & n.a. & n.a. \\
Amputation (Binary) & 2 & $0.99\pm0.04$ & 0.989 & n.a. & n.a. & n.a. \\
\hline
\end{tabular*}
\end{table}

\subsubsection{Held-Out IMIS Database}
This analysis was an ad-hoc test of whether the classification networks, trained exclusively on the SIR corpus, can extrapolate to an independent database with a different composition. The evaluation set is an inference sample of 1,197 OSHA fatality narratives drawn from the Integrated Management Information System (IMIS) database, none of which the networks were trained on. Several major differences between the datasets resulted in interesting effects. This dataset includes fatalities whereas SIR does not, so the injury outcomes and the language describing them differ systematically from anything the models have seen prior. Additionally, the set carries no gold OIICS labels which governs any claims in these results as these accidents were not coded directly to the same schema. Accuracy therefore cannot be measured, and performance herein means human judgement and agreement with a small set of author-assigned labels, not a measured error rate.

A random sample of 100 records (seed 42) was drawn for close inspection. One of the authors labeled these records blindly before any network or other output was analyzed. The neural network (MLP) predictions were then compared with an ensemble of LLMs using the majority vote classification procedure established in prior work \cite{Sme26}. Reasoning was disabled for all capable models and temperature fixed at 0.1 during generation. All agreement figures in \hyperref[tab:imis]{Table \ref{tab:imis}} are computed relative to the author labels for the 100-record sample.

\begin{table}[ht]
\centering
\caption{Random sample agreement over held-out incidents from IMIS ($n=100$)}
\label{tab:imis}
\begin{tabular*}{\textwidth}{@{\extracolsep{\fill}} l *{5}{c} @{}}
\hline
\textbf{OIICS Category} & \textbf{MLP} & \textbf{gpt-4o-mini} & \textbf{gpt-5-mini} & \textbf{gemini-2.5-flash-lite} & \textbf{gemini-3.5-flash-lite} \\
\hline
Event & 54.0 & 57.0 & 56.0 & 43.0 & \textbf{64.0} \\
Source & \textbf{62.0} & 6.00 & 20.0 & 36.0 & 44.0 \\
Nature & \textbf{60.0} & 37.0 & 42.0 & 35.0 & 46.0 \\
Part of Body & \textbf{69.0} & 19.0 & 29.0 & 28.0 & 36.0 \\
Hospitalized (Binary) & 37.0 & 37.0 & 44.0 & \textbf{83.0} & 82.0 \\
Amputation (Binary) & 61.0 & 21.0 & 96.0 & 92.0 & \textbf{100} \\
\hline
Mean & 57.2 & 29.5 & 47.8 & 52.8 & \textbf{62.0} \\
\end{tabular*}
\end{table}

The neural networks are competitive with frontier LLMs, posting a mean agreement of 57.2\% across the six OIICS fields, second only to gemini-3.5-flash-lite at 62.0\%. The four multiclass MLPs land at roughly 60-70\% agreement (Source 62\%, Part of Body 69\% strongest), which is plausible with real OIICS coding performance and broadly consistent with the $\approx 75\%$ held-out accuracy observed during SIR testing. On the other hand, the two binary fields are barely informative and should be down weighted or dropped entirely. In the author-labeled set there were no positive cases in this sample, so the metric measures only false-positive rate, and a model that always predicts ``no'' (gemini-3.5) scores a perfect 100\% without exhibiting any detection ability. The MLP, by contrast, predicted amputation for 39 of 100 fatalities, yielding 39 false positives and no recoverable true positives. The predictions for Hospitalized are also degenerate, emitting ``hospitalized'' for 99 of 100 cases, though this reading is sensitive to how IMIS actually defines the field. Either way, a single class prediction for essentially every record carries no information, and this reinforces that a dedicated neural network may be unnecessary for binary flags.

\subsubsection{Quality Scoring Mechanism}
Each prediction carries not only a top-ranked OIICS code but, but also a measure of how confident the model is in its prediction, termed its determinacy ($det$) in \hyperref[eq:determinacy]{Eq. \ref{eq:determinacy}}. Intuitively, a high determinacy means the model committed to one code, while a low determinacy means it spread its prediction across several plausible indices and is effectively undecided. Each record receives a single composite quality score, computed as the mean determinacy across the four multiclass columns (Event, Source, Nature, and Part of Body). The two binary fields, Hospitalized and Amputation, are deliberately excluded from this fused score. The previous held-out analysis established that both binaries are degenerate on this data, so their determinacy is high but uninformative. It is worth noting that each category contains a designated dustbin code for narratives that cannot be assigned to a class, each mapped to the ``9999 -- Nonclassifiable'' designation in the source database. Since a prediction into one of these dustbin classes conveys no real classification, its determinacy is explicitly set to 0 for that column regardless of how confidently the model predicted it.

Under this scoring mechanism, SIR appears to have a balanced distribution (\hyperref[tab:quality]{Table \ref{tab:quality}}). Determinacy is moderate and even across all four columns (0.78-0.85), and only 20\% of records saturate above 0.9. This is the signature of a useful quality evaluation, or one that is actually spreading records apart rather than declaring nearly everything as confident. Alternatively, IMIS is more saturated. Although its quality mean (0.811) is nearly identical to SIR's (0.812), the underlying meaning is less intuitive. Event and Source determinacy are pinned near 1.0 (0.976 and 0.987), Nature has collapsed to 0.456 as its heavy ``Nonclassifiable'' dustbin category selection brings determinacy down, and 42\% of records saturate at $\geq$ 0.9. The 100-record random sample tracks the full 1,197 record IMIS set closely on every metric.

\begin{table}[ht]
\centering
\caption{Quality scoring statistics for test and held-out datasets}
\label{tab:quality}
\begin{tabular*}{\textwidth}{@{\extracolsep{\fill}} l c c c @{}}
\hline
\textbf{Metric} & \textbf{SIR (test)} & \textbf{IMIS (237310)} & \textbf{IMIS (sample)} \\
\hline
Records & 15,439 & 1,197 & 100 \\
Quality mean & 0.812 & 0.811 & 0.802 \\
Quality median & 0.832 & 0.750 & 0.750 \\
Quality std. dev. & 0.112 & 0.171 & 0.171 \\
Quality min / max & 0.000 / 0.997 & 0.233 / 1.00 & 0.436 / 1.00 \\
\% scoring $\geq$ 0.90 & 20\% & 42\% & 40\% \\
det -- Event & 0.816 & 0.976 & 0.981 \\
det -- Source & 0.778 & 0.987 & 0.976 \\
det -- Nature & 0.851 & 0.456 & 0.412 \\
det -- Part of Body & 0.802 & 0.824 & 0.837 \\
\hline
\end{tabular*}
\end{table}

\subsubsection{Word Saliency Example}
\begin{figure}[ht]
\centering
\includegraphics[width=0.99\textwidth]{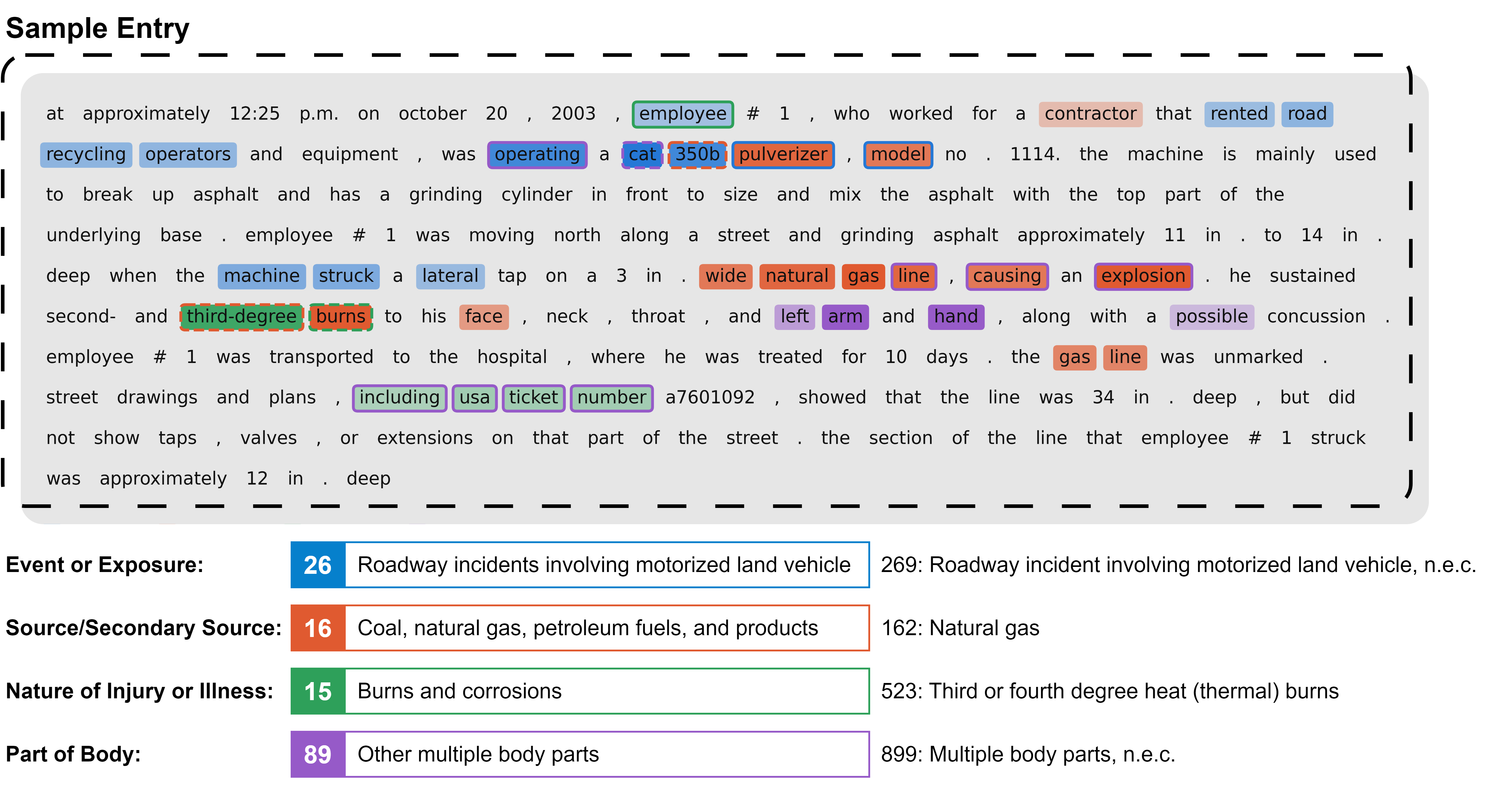}
\caption{Example saliency output and OIICS predictions for accident narrative from IMIS}
\label{fig:saliency}
\end{figure}

To make an individual classification more interpretable, word-level saliency was computed by a leave-one-out (LOO) principle as described by \hyperref[eq:loo]{Eq. \ref{eq:loo}}. Each word was removed from the narrative in turn, the modified text is re-embedded and re-scored, and the resulting drop in the predicted probability of a categories target code measures that words contribution. Words whose removal most reduces the model's confidence are the most salient, and in \hyperref[fig:saliency]{Figure \ref{fig:saliency}} these are highlighted per column, color-matched to the four OIICS structures. The border around each word, which may be color-coordinated with another column, represent overlapping influence on several categories. A solid border in this case represents a two-way overlap, and a dashed border represents the words influence on two or more categories.

\subsection{Information Retrieval Tasks}
RAG in its naïve form is a two step pattern. A corpus of documents is embedded once into a shared vector space, and a query embedded with the same model and compared against every stored document by cosine similarity (\hyperref[eq:cosine]{Eq. \ref{eq:cosine}}). In the following results, the same naïve RAG mechanism is applied three distinct retrieval tasks, each defined by what serves as the query and what is being ranked: (1) historical accident retrieval, in which a construction activity description queries a corpus of labeled OSHA accident narratives to surface comparable past incidents for a JSA; (2) relevant imagery retrieval, linking planned work to visual references of the associated hazards and controls; and (3) industry document QA, in which a planner's question queries a curated corpus of best practice guides, agency standards, and equipment manuals to retrieve supplementary guidance. Each task reuses the identical similarity procedure and is evaluated with the same information retrieval metrics, differing only in the query and corpus involved.

\subsubsection{Historical Accidents}
As described in the methodology, the first retrieval tasks drew on past incident narratives from databases to surface how similar work has led to injury in the field. Building on the prior analysis of PennDOT Pub 517 \cite{Sme26}, in which both SIR and IMIS narratives were labeled to their respective construction activities, a total of 965 narratives were mapped to 46 activities described in the manual. Reported metrics in \hyperref[tab:accidents]{Table \ref{tab:accidents}} are macro-averaged over the activities, so each contributes equally regardless of how many accidents they carry. As a result, an activity mapped to only one single narrative, such as Pipe Flushing, and activity with 82 relevant narratives weigh the same in the mean -- which indicate inherent limitations of this evaluation. \hyperref[tab:accidents]{Table \ref{tab:accidents}} reports the averaged retrieval statistics across five embedding models. The two proprietary OpenAI models lead clearly, with text-embedding-3-large (MAP 0.277) strongest on every metric and text-embedding-3-small (0.217) second. The three open models form a tight cluster below them: Qwen-Embedding-0.6b (0.175), nomic-embed-text-v1.5 (0.173), and all-MiniLM-L6-v2 (0.161).

\begin{table}[ht]
\centering
\caption{Averaged statistics for incident retrieval over 46 job activities}
\label{tab:accidents}
\begin{tabular*}{\textwidth}{@{\extracolsep{\fill}} l c c c c c @{}}
\hline
\textbf{Model} & \textbf{MAP} & \textbf{MRR} & \textbf{NDGC@10} & \textbf{Recall@10} & \textbf{Recall@100} \\
\hline
text-embedding-3-large & \textbf{0.277} & \textbf{0.525} & \textbf{0.357} & \textbf{0.231} & \textbf{0.615} \\
text-embedding-3-small & 0.217 & 0.456 & 0.300 & 0.192 & 0.568 \\
Qwen-Embedding-0.6b & 0.175 & 0.442 & 0.286 & 0.138 & 0.489 \\
nomic-embed-text-v1.5 & 0.173 & 0.412 & 0.253 & 0.155 & 0.470 \\
all-MiniLM-L6-v2 & 0.161 & 0.374 & 0.242 & 0.172 & 0.462 \\
\hline
\end{tabular*}
\end{table}

Performance was uneven across activities, and inspecting the per-activity average precision for the best model (text-embedding-3-large) makes the pattern clear. The easiest activities to retrieve are distinct and usually tied to specialized equipment: Wood-Chipper -- Chipping Brush ($n=2, \text{AP}=1.00$); Broom/Towable Broom ($n=2, \text{AP}=0.833$); Skid Steer -- General ($n=28, \text{AP}=0.670$); Scaffold Erecting/Dismantling ($n=26, \text{AP}=0.619$); Garage -- Tire Removal/Install ($n=4, \text{AP}=0.604$). On the other hand, the hardest to retrieve are generic pavement maintenance procedures whose narratives share vocabulary and context with neighboring activities: Airless Paint Truck ($n=12, \text{AP}=0.027$); Patching -- General ($n=27, \text{AP}=0.026$); Shoulder Cutting -- General ($n=11, \text{AP}=0.025$); Work Zone Traffic Control -- Mobile Setup ($n=9, \text{AP}=0.020$); Pipe Flushing -- General ($n=1, \text{AP}=0.012$).

\subsubsection{Related Imagery}
\begin{figure}[ht]
    \centering
    \begin{minipage}{0.24\textwidth}
        \centering
        \textbf{0.74} \\[3pt]
        \includegraphics[width=\textwidth]{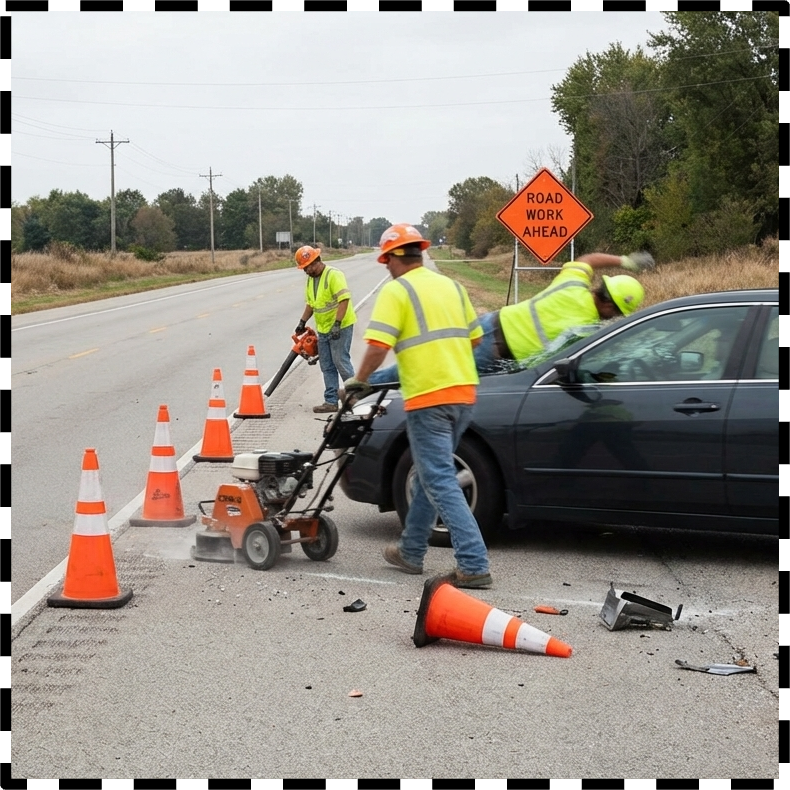}
    \end{minipage}\hfill
    \begin{minipage}{0.24\textwidth}
        \centering
        \textbf{0.71} \\[3pt]
        \includegraphics[width=\textwidth]{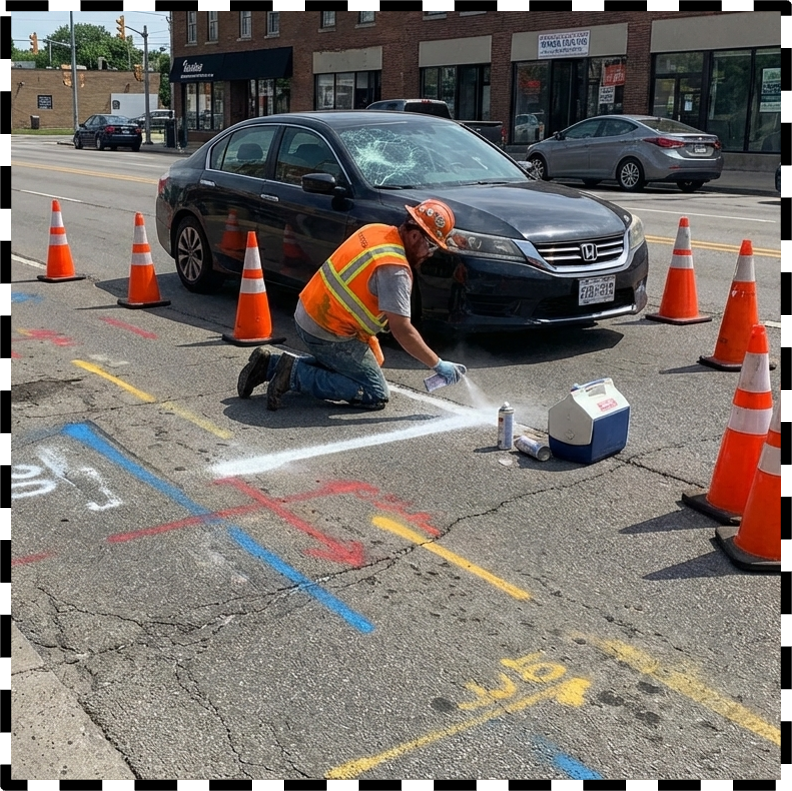}
    \end{minipage}\hfill
    \begin{minipage}{0.24\textwidth}
        \centering
        \textbf{0.69} \\[3pt]
        \includegraphics[width=\textwidth]{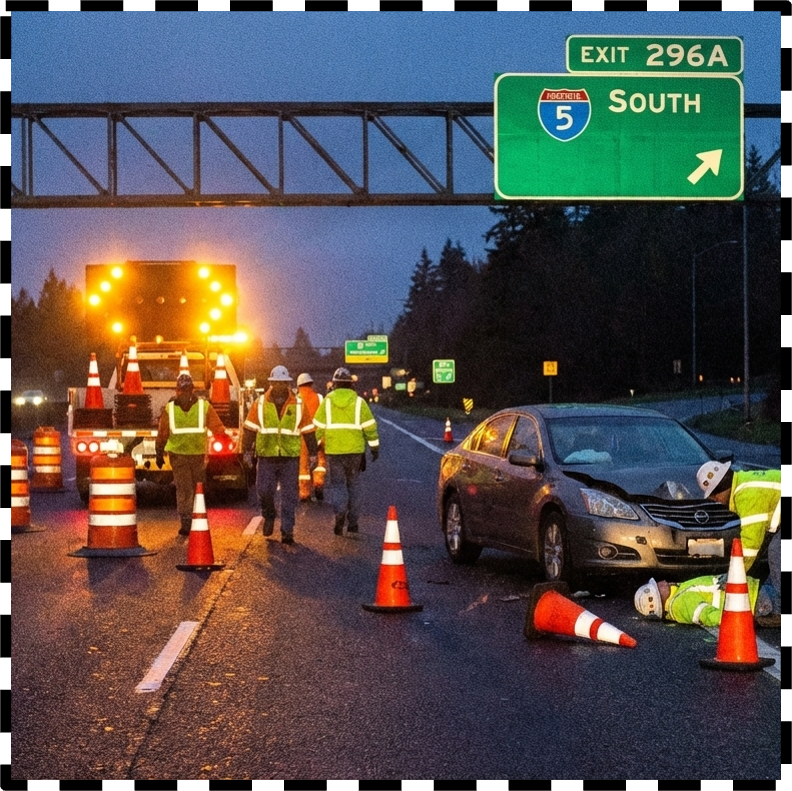}
    \end{minipage}\hfill
    \begin{minipage}{0.24\textwidth}
        \centering
        \textbf{0.67} \\[3pt]
        \includegraphics[width=\textwidth]{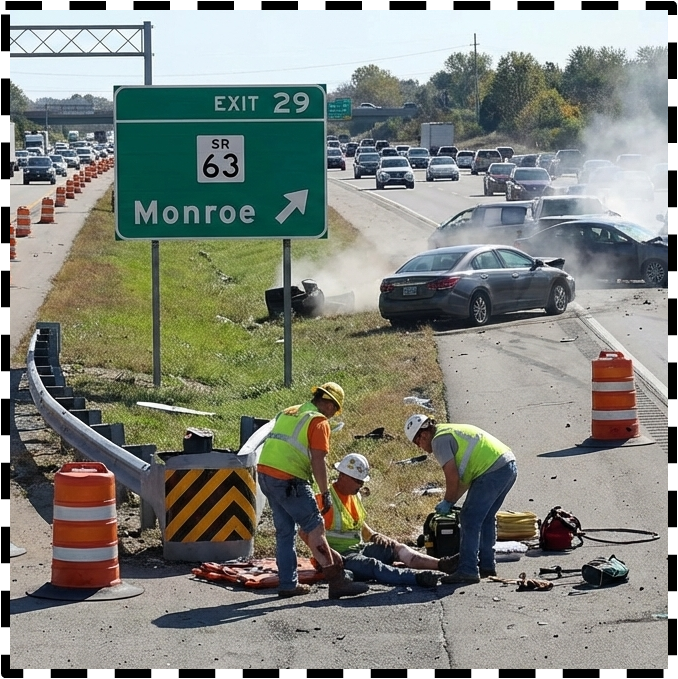}
    \end{minipage}
    \caption{Retrieved imagery for vehicle intrusion accidents based on similarity score}
    \label{fig:imagery}
\end{figure}

The next retrieval task linked a planned activity to reference images of the associated hazards and work conditions, giving crews a visual aid to accompany the accident narratives in a safety briefing. In its current form the images were retrieved indirectly, which is a notable limitation in this study. Each image was paired with the accident narrative that generated it, according to the methodology described in \cite{Nee26}. Retrieval was performed over those coupled narratives rather than the image content itself, so the ranking metrics here are identical to the historical accident task and the image is strictly informative to the practitioner. \hyperref[fig:imagery]{Figure \ref{fig:imagery}} shows an example for vehicle intrusion incidents, where the narrative for record 95 of SIR (a worker grinding pavement markings struck by a vehicle that drove through the coned off zone) surfaces images depicting that same struck-by scenario.

\subsubsection{Industry Document Question Answering}
The final retrieval task evaluated whether a planner's question can surface the correct passage from the curated corpus of parsed industry documents. Using the synthetic question generation pipeline described earlier \cite{NVI26}, 6,214 query-passage pairs were generated with gpt-4o-mini. \hyperref[tab:industry]{Table \ref{tab:industry}} reports the results across the same five embedding models as the previous tasks. The authors did not manually vet the quality of the synthetic questions themselves, which should be explored in future studies. The table therefore indicates when retrieval works rather than certifying the questions. The strongest model correctly places the target passage first only about 45\% of the time (Acc@1) and its top-1 relevance is still roughly a coin flip. Retrieval improves substantially with depth (Acc@10 reaching 0.79), but pinpoint accuracy remains limited.

The most striking result is that the open-weight Qwen-Embedding-0.6b model leads to the evaluation surpassing the proprietary OpenAI models on all metrics. This reverses the pattern seen in the historical accident task, where the commercial models were clearly strongest, and it shows that a capable open decoder-based embedding model is fully within reach of frontier general purpose models on document QA; however, this capability comes at a cost. Qwen has substantially higher inference and GPU requirements, whereas the lighter encoder only are much faster to run and can be inferred entirely on consumer grade CPUs. all-MiniLM-L6-v2 in particular is the weakest performer, which is partly attributable to its small context window truncating document segments.

\begin{table}[ht]
\centering
\caption{Retrieval metrics for parsed industry documents and synthetic questions ($n=6,214$)}
\label{tab:industry}
\begin{tabular*}{\textwidth}{@{\extracolsep{\fill}} l c c c c c @{}}
\hline
\textbf{Model} & \textbf{Acc@1} & \textbf{Acc@10} & \textbf{MRR@10} & \textbf{NDGC@10} & \textbf{MAP@100} \\
\hline
text-embedding-3-large & 0.334 & 0.733 & 0.459 & 0.525 & 0.468 \\
text-embedding-3-small & 0.358 & 0.758 & 0.484 & 0.550 & 0.492 \\
Qwen-Embedding-0.6b & \textbf{0.451} & \textbf{0.789} & \textbf{0.562} & \textbf{0.617} & \textbf{0.562} \\
nomic-embed-text-v1.5 & 0.320 & 0.647 & 0.421 & 0.475 & 0.430 \\
all-MiniLM-L6-v2 & 0.294 & 0.605 & 0.389 & 0.441 & 0.399 \\
\hline
\end{tabular*}
\end{table}

\section{Discussion}
\subsection{Incident Classification and Scoring}
The classification models perform well across the OIICS structures, but their performance is best understood as a function of task difficulty rather than as a single accuracy. The confidence distributions in \hyperref[fig:oiics]{Figure \ref{fig:oiics}} appear to be as informative as the metrics in \hyperref[tab:oiics]{Table \ref{tab:oiics}}. An example makes the spread tangible. Consider a fatal paving machine incident in which a screed operator lost his balance, fell, and was struck by the machine while it backed during paving. The Amputation model outputs 0.99 for ``no amputation,'' which is an easy high confidence call that lands in the narrow spike near 1.0. By contrast, the Source model must choose from 79 codes, and a paving machine sits ambiguously between a vehicle and construction/process machinery. The MLP may predict roughly 0.40 on one equipment code, 0.30 on another, and scatter the remainder across neighbors. Its top prediction therefore carries only $\approx 0.40$ confidence and falls in the broad left shoulder of the Source curve. Operationally, the top-1 label should not always be considered the underlying truth; however, the correct code is very likely somewhere in the model's short list. This behavior explains why Source recovers from an Acc@1 of 0.694 to an Acc@3 of 0.876. A spread out prediction should be read not as a committed single label but as a ranked set of candidate codes for human confirmation.

The encouraging pattern is that the reported probability is a usable, roughly calibrated signal of when a label can be trusted and when it should instead be surfaced as a top-3 shortlist for a human reviewer. Overall, the Source model is the weakest on every statistic, which is unsurprising given that it resolves the largest and most overlapping taxonomy. A larger number of codes contrasts with the stronger Part of Body (Acc@1 = 0.843) and Nature (Acc@1 = 0.815) over 46 and 41 classes, respectively. This difficulty is the primary reason classification was performed at OIICS Level 2, the ``major groups'' level, as lower levels of the hierarchy expand quickly into hundreds of fine-grained classes (upwards of 400). A network attempting to resolve such large and convoluted label spaces is prone to becoming trapped in local minima early in training and yields little practical benefit. It is suggested to leave lower level, more granular coding to future work. Moreover, it should also be noted that no grid search was conducted over network hyperparameters, so the reported figures reflect a nonexhaustively optimized configuration.

A useful byproduct of this approach is cross schema conversion. Since the models learn to map free-text narratives onto coded structures, they enable translating incidents recorded under one classification system (e.g., IMIS) into another (SIR/OIICS) with reasonable accuracy. The four multiclass columns transfer at roughly 60-70\% agreement with author labels, consistent with the $\approx 75\%$ held-out testing accuracy. 

Beyond classification performance, a scoring mechanism was proposed to fuse the inference of several independent MLP classifiers into a single, quantified estimate of how well a narrative can be coded into the OIICS schema (a proxy to narrative quality). Rather than treating each OIICS column in isolation, the system reads the determinacy of every prediction and combines them into one composite quality score. This is an appealing property, as it turns six separate classification acts into a single interpretable number that a reporter or reviewer can act on without reading any of the underlying code assignments. On SIR, this premise largely holds and the score behaves as a trustworthy, discriminating signal (\hyperref[tab:quality]{Table \ref{tab:quality}}). Event and Source are the healthiest networks, while Nature and Part of Body are compromised by heavy routing into ``nonclassifiable'' categories. The remaining binary categories, Hospitalized and Amputation are unfortunately degenerate, they do not provide an adequate signal for the held-out dataset (\hyperref[tab:imis]{Table \ref{tab:imis}}). This reinforces that a dedicated neural network may be unnecessary for the binary flags. Some of this behavior may be attributable to IMIS narratives being more detailed and occasionally more convoluted than SIR, which stress tests the classifiers more rigorously.

The takeaway is that fusing multiple MLP inferences into a single score is a sound and valuable concept. However, as currently constructed, it primarily measures softmax saturation instead of true input quality. Making it robust across databases will require further calibration of and potentially a labeled sample to validate the score's core premise that determinacy correlates with correctness. The pseudo-quality score is instead presented as a preliminary metric that warrants deeper analysis in subsequent studies, and it is recommended that the system surface the top-3 candidate predictions per column and leave final selection to the end user rather than committing to a single automated label.

Finally, to steer away from the black box nature of machine learning and LLMs, the leave-one-out saliency mechanism can be a useful tool in evaluating model understanding. The example in \hyperref[fig:saliency]{Figure \ref{fig:saliency}} illustrates that the networks key on semantic evidence rather than surface cues. The Event model's prediction of 26: ``Roadway incidents involving motorized land vehicle'' is driven by words such as ``machine,'' ``struck,'' and the equipment terms describing the asphalt pulverizer, tying the classification to the vehicular strike at the center of the narrative. Source (16: ``Coal, natural gas, petroleum fuels, and products'') is anchored on ``natural gas line'' and ``explosion.'' Nature (15: ``Burns and corrosions'') focuses on ``third-degree'' and ``burns,'' and Part of Body (89: ``Other multiple body parts'') on the enumerated injuries to ``face,'' ``arm,'' and ``hand.'' Each column attends to the span of text that actually motivates its label, and the fact that the four models fixate on largely disjoint, contextually correct phrases indicates the predictions reflect genuine reading of the narrative rather than a shared lexical shortcut.

\subsection{Information Retrieval Tasks}
The accident retrieval results (\hyperref[tab:accidents]{Table \ref{tab:accidents}}) establish that surfacing historical incidents via retrieval is feasible. Since relevance is single label, and several hard activities are procedures whose accidents plausibly belong to more than one activity (e.g., Patching and Shoulder Cutting), the reported figures are best read as lower bounds on for retrievability rather than as absolute ceilings. Nonetheless, the best model (text-embedding-3-large) gets at least one relevant accident into the top 100 for almost every activity (recall@100 $\approx 0.62$), which is excellent for a ``show me comparable incidents'' aid that a person reviews. However, at the shallow depths, a reviewer can expect only about a quarter of the relevant accidents (recall@10 $\approx 0.23$), so skimming the top ten will miss some applicable events. Performance also depends heavily on the activity: distinctive, equipment-specific tasks retrieve very well, whereas generic pavement maintenance jobs whose narratives blur into neighboring activities retrieve poorly. For supplementing a safety briefing with real, comparable accidents, off-the-shelf embeddings are dependable enough to be genuinely useful -- provided a human stays in the loop and the system presents a ranked set of candidates rather than committing to a single match.

Several known limitations bound these results and point toward future research directions. Each activity was represented by a single free-text description, so retrieval is sensitive to query wording. This study did not explore variations in the input activity descriptions or their effect on retrieval. Off-the-shelf embeddings suffice to establish the result and to power a lenient retrieval aid, but they may significantly benefit from additional fine-tuning. Such adaptation would also enable the framework's call for continuous improvement efforts. An in-house repository of near-misses and contractor-specific incidents could be folded into the searchable corpus, but realizing its value at the shallow depths that crews may actually consult will require models tuned for this retrieval task rather than used out of the box.

For the second image retrieval task, it is evident that presenting JSA documents more effectively argues for pairing them with visuals. However, this technique is hindered by a data availability gap where image datasets of real-world construction hazards are difficult to collect and are presently sparse \cite{Guo21}. Capturing such imagery directly raises site access limitations, ethical concerns around documenting unsafe practice before intervening, and companies' reluctance to release these images publicly. For this reason, the images used in the present prototype are entirely AI-generated to demonstrate the concept, and in any deployment they should be replaced with an organization's own proprietary imagery where it exists. A further discrepancy worth flagging is that similarity is currently computed on the text rather than the picture, so a retrieved image is only as relevant as the narrative attached to it. Genuine image-based retrieval would require a multimodal embedding model and further tuning, and cosine similarity over text alone is unlikely to be sufficient. Even with these caveats, the core value is immediate -- putting a realistic reference image of the hazard in front of employees in the field, rather than a generic illustration -- and it establishes a visual retrieval stream that a proprietary image library and a multimodal encoder could make fully operational in future work.

The last industry document QA retrieval task produced the study's most surprising result: the open-weight decoder-based Qwen-Embedding-0.6b model led every proprietary and general purpose model tested, surpassing the OpenAI embeddings all metrics (\hyperref[tab:industry]{Table \ref{tab:industry}}). This reverses the ordering seen in the accident retrieval task, where the commercial models were clearly strongest, and it demonstrates that a capable open embedding model is not merely within reach of frontier proprietary models on this workload but can exceed them. That advantage, however, comes with a practical tradeoff that matters for a framework built on local tooling. Qwen's stronger retrieval is paid for in substantially higher inference cost and GPU requirements, whereas lighter encoders run much faster on modest hardware. all-MiniLM-L6-v2 in particular was the weakest performer, in part because its small context window truncates longer document segments before they are fully embedded, yet it remains by far the smallest and fastest option.

These results suggest that document QA task would most likely to benefit from targeted domain fine-tuning in future work. NVIDIA's synthetic data generation and fine-tuning pipeline is a promising route to that adaptation, yet its effectiveness is heavily dependent on the parsing and chunking strategy applied to the source documents. It also would likely require a more capable model than gpt-4o-mini for question generation and LLM-as-a-judge evaluation to yield training and test data trustworthy enough to justify the fine-tuning. Until such adaptation is carried out and validated, the industry document QA stream should be regarded as a functional but preliminary capability that is useful for surfacing candidate passages under human review.

\section{Conclusions}
This study introduced and evaluated an AI Safety Assistant (AISA) framework for highway construction, built around the continuous improvement of Job Safety Analysis (JSA) reports and grounded in deterministic, local inference rather than reliance on cloud-based frontier models. A prototype has been developed for near-future field exposure, in which industry partners and safety professionals can assess the utility of the proposed features directly. Across classification, quality scoring task, and three retrieval streams (historical accidents, related imagery, and industry document QA), the results support the framework's central premise: that low cost, auditable, offline tooling can meaningfully surface the lessons of past work during the activity planning stage.

The classification results demonstrate the utility of lightweight multilayer perceptron (MLP) models for deterministic coding of accident narratives into the Occupational Injury and Illness Classification System (OIICS). Trained against the Severe Injury Reports (SIR) database and evaluated on held-out data, the multiclass networks performed strongest where the coding structure was clearly separated (e.g., Part of Body and Nature of Injury) and weakest on the largest, most overlapping label spaces (Source of Injury). A useful byproduct is the ability to translate incidents recorded under one repository into another with reasonable accuracy, as demonstrated by coding an external database into the SIR/OIICS schema. Since these models run offline and predict reproducible outputs, they are well suited to applications that require grounded, deterministic inference rather than the probabilistic behavior of large generative models.

The prototype scoring mechanism, which fuses the predictions of several MLP inferences into a single quantified estimate of a narrative's pseudo-quality, is presented as a promising but preliminary contribution. It behaves as a trustworthy signal on in-distribution data, but analysis of out-of-distribution fatalities exposed the ways it can distort the networks confidence. Alongside this, the leave-one-out saliency analysis surfaces which words influence each prediction, offering a step toward interpretable AI by letting a reviewer see the textual evidence behind a code rather than accepting it as a black-box label. Both mechanisms warrant deeper study and labeled samples to validate the scores premise.

For the retrieval tasks, a benchmark of activity-to-accident matching across six embedding models was compiled. The results demonstrate that dense retrieval recovers relevant historical accidents above chance for the majority of activities, but they require a still human reviewer to select the most relevant. Although results vary across the three tasks, accident retrieval favored proprietary embeddings while industry document QA was led by an open-weight decoder-based model. It is evident that recent advances in open-weight embedding models proved capable of reliably drawing information from external sources, which is essential to a privacy-preserving, local first design. Notably, this study deliberately examined only naive RAG. It is anticipated that more advanced retrieval methods, such as reranking, hybrid, and multihop retrieval, would only improve performance.

Several limitations bound these conclusions and define the studies' conceptual nature. By design, the methods only use models out of the box without fine-tuning, and no ablation studies were conducted over embedding model types for OIICS classification. In the current state, true multimodal capability remains out of reach for the edge devices and lightweight models this framework targets, so image retrieval currently operates over coupled narratives rather than image content, and genuine visual understanding would likely require fine-tuned multimodal encoders. The imagery in this prototype is AI-generated for demonstration and should be replaced with proprietary images in deployment. Eventually, visuals may be crafted on the fly to support JSA and daily toolbox talks. Human feedback and iteration are still required, and the framework should be regarded as proof of concept until validated further.

Ultimately, the results of this study support the capability of the proposed AISA framework in highway construction scenarios grounded in historical record, and for potential agentic applications. Its design deliberately favors local models for the privacy and security of sensitive and proprietary data, folding new accidents and near-misses back into the retrieval corpus so that each round of safety planning learns from the last. Beyond standalone use, the deterministic components developed here are intended to serve as reliable building blocks for future agentic applications, which could be extended with sub-agent architectures and additional external data sources (e.g., weather, construction scheduling). In providing verifiable, offline tooling this framework complements rather than replaces the agentic LLM pipelines now entering practice.

\section*{Acknowledgements}
The authors would like to acknowledge the Anthony Gill Chair and Impactful Resilient Infrastructure Science and Engineering (IRISE) Consortium for funding this project. We are indebted for the advice and assistance provided by the following representatives of IRISE member organizations that comprised the technical panel who guided work on the project: Pennsylvania Turnpike Commission, Pennsylvania Department of Transportation, Contractors Association of Western Pennsylvania, Michael Baker International, Allegheny County, Golden Triangle Construction, and the Federal Highway Administration.

\section*{Disclaimer: On the use of Generative AI}
OpenAI's ChatGPT was used to assist proofreading and revision of the manuscript. The authors of this paper acknowledge the limitations of LLMs, such as potential biases, errors, and gaps in knowledge.

\section*{Author Contributions}
The authors confirm contribution to the paper as follows: study conception and design: M. Smetana, L. Khazanovich; data collection: M. Smetana, T. Neece; analysis and interpretation of results: M. Smetana, L. Khazanovich. T. Neece; draft manuscript preparation: M. Smetana. L. Khazanovich, T. Neece. All authors reviewed the results and approved the final version of the manuscript.

\bibliographystyle{unsrtnat}
\bibliography{references}

\end{document}